\documentclass[runningheads]{llncs}
\usepackage{graphicx}
\usepackage{xcolor}
\usepackage[colorlinks=true,citecolor=blue,urlcolor=blue]{hyperref}
\usepackage{array,booktabs}
\usepackage{algorithm,algpseudocode}
\usepackage{amsfonts}
\usepackage{amsmath}
\usepackage{comment}
\usepackage{color}
\usepackage[misc]{ifsym}
\usepackage{bbding}
\def\eg{\textit{e.g.}}
\def\ie{\textit{i.e.}}
\def\etal{\textit{et al.}}

\begin{document}
%
\title{Shape-aware Meta-learning for
Generalizing \\ Prostate MRI Segmentation to Unseen Domains}
\titlerunning{Shape-aware Meta-learning for Domain Generalization}
%
\author{Quande Liu\inst{1} \and Qi Dou \inst{1,2}\Envelope \and PhengAnn Heng\inst{1,3}}
\institute{Department of Computer Science and Engineering, The Chinese University of Hong Kong, Hong Kong SAR, China\\\email{\{qdliu, qdou, pheng\}@cse.cuhk.edu.hk} \and T Stone Robotics Institute, The Chinese University of Hong Kong, Hong Kong SAR, China \and Guangdong Provincial Key Laboratory of Computer Vision and Virtual Reality Technology, Shenzhen Institutes of Advanced Technology, Chinese Academy of Sciences, Shenzhen, China}%


%
%
\maketitle              
\begin{abstract}
Model generalization capacity at domain shift (\eg, various imaging protocols and scanners) is crucial for deep learning methods in real-world clinical deployment.
%
This paper tackles the challenging problem of domain generalization,~\ie, learning a model from multi-domain source data such that it can directly generalize to an unseen target domain. We present a novel shape-aware meta-learning scheme to improve the model generalization in prostate MRI segmentation. 
Our learning scheme roots in the gradient-based meta-learning,
by explicitly simulating domain shift with virtual meta-train and meta-test during training. 
Importantly, considering the deficiencies encountered when applying a segmentation model to unseen domains (i.e., incomplete shape and ambiguous boundary of the prediction masks), we further introduce two complementary loss objectives to enhance the meta-optimization, by particularly encouraging the \emph{shape compactness} and \emph{shape smoothness} of the segmentations under simulated domain shift. 
We evaluate our method on prostate MRI data from six different institutions with distribution shifts acquired from public datasets.
Experimental results show that our approach outperforms many state-of-the-art generalization methods consistently across all six settings of unseen domains\footnote{Code and dataset are available at \url{https://github.com/liuquande/SAML}}.


\end{abstract}

\section{Introduction}
Deep learning methods have shown remarkable achievement in automated medical image segmentation~\cite{jia2019hd,milletari2016v,zhu2019boundary}. 
However, the clinical deployment of existing models still suffer from the performance degradation under the distribution shifts across different clinical sites using various imaging protocols or scanner vendors. 
Recently, many domain adaptation~\cite{chen2020unsupervised,kamnitsas2017unsupervised} and transfer learning methods~\cite{gibson2018inter,karani2018lifelong} have been proposed to address this issue, while all of them require images from the target domain (labelled or unlabelled) for model re-training to some extent. 
In real-world situations, it would be time-consuming even impractical to collect data from each coming new target domain to adapt the model before deployment. 
Instead, learning a model from multiple source domains in a way such that it can directly generalize to an unseen target domain is of significant practical value. %
This challenging problem setting is \emph{domain generalization (DG)}, in which no prior knowledge from the unseen target domain is available during training.
%
%

%

Among previous efforts towards the generalization problem~\cite{gibson2018inter,liu2020ms,yoon2019generalizable}, a naive practice of aggregating data from all source domains for training a deep model (called `DeepAll' method) can already produce decent results serving as a strong baseline. It has also been widely used and validated in existing literature~\cite{chen2019improving,dou2020unpaired,yao2019strong}. 
%
On top of DeepAll training, several studies added data augmentation techniques to improve the model generalization capability~\cite{zhang2020generalizing,paschali2018generalizability}, assuming that the domain shift could be simulated by conducting extensive transformations to data of source domains. Performance improvements have been obtained on tasks of cardic~\cite{chen2019improving}, prostate~\cite{zhang2020generalizing} and brain~\cite{paschali2018generalizability} MRI image segmentations, yet the choices of augmentation schemes tend to be tedious with task-dependence. 
Some other approaches have developed new network architectures to handle domain discrepancy~\cite{kouw2019cross,yang2018generalizing}. 
Kour~\etal~\cite{kouw2019cross} developed an unsupervised bayesian model to interpret the tissue information prior for the generalization in brain tissue segmentation. 
A set of approaches~\cite{aslani2020scanner,otalora2019staining} also tried to learn domain invariant representations with feature space regularization by developing adversarial neural networks.
Although achieving promising progress, these methods rely on network designs, which introduces extra parameters thus complicating the pure task model.

Model-agnostic meta-learning~\cite{finn2017model} is a recently proposed method for fast deep model adaptation, which has been successfully applied to address the domain generalization problem~\cite{balaji2018metareg,dou2019domain,li2018learning}.
The meta-learning strategy is flexible with independence from the base network, as it fully makes use of the gradient descent process. However, existing DG methods mainly tackle image-level classification tasks with natural images,
which are not suitable for the image segmentation task that requires pixel-wise dense predictions.
An outstanding issue remaining to be explored is how to incorporate the shape-based regularization for the segmentation mask during learning, which is a distinctive point for medical image segmentation.
In this regard, we aim to build on the advantages of gradient-based meta-learning, while further integrate shape-relevant characteristics to advance model generalization performance on unseen domains.

We present a novel \textbf{s}hape-\textbf{a}ware \textbf{m}eta-\textbf{l}earning (SAML) scheme for domain generalization on medical image segmentation.
Our method roots in the meta-learning episodic training strategy, to promote robust optimization by simulating the domain shift with meta-train and meta-test sets during model training. 
Importantly, to address the specific deficiencies encountered when applying a learned segmentation model to unseen domains (i.e., incomplete shape and ambiguous boundary of the predictions), we further propose two complementary shape-aware loss functions to regularize the meta optimization process. 
First, we regularize the \emph{shape compactness} of predictions for meta-test data, enforcing the model to well preserve the complete shape of segmentation masks in unseen domains. Second, we enhance the \emph{shape smoothness} at boundary under domain shift, for which we design a novel objective to encourage domain-invariant contour embeddings in the latent space. We have extensively evaluated our method with the application of prostate MRI segmentation, using public data acquired from six different institutions with various imaging scanners and protocols. Experimental results validate that our approach outperforms many state-of-the-art methods on the challenging problem of domain generalization, as well as achieving consistent improvements for the prostate segmentation performance across all the six settings of unseen domains.





\section{Method}
\begin{figure*}[t]
	\centering
	\includegraphics[width=\textwidth]{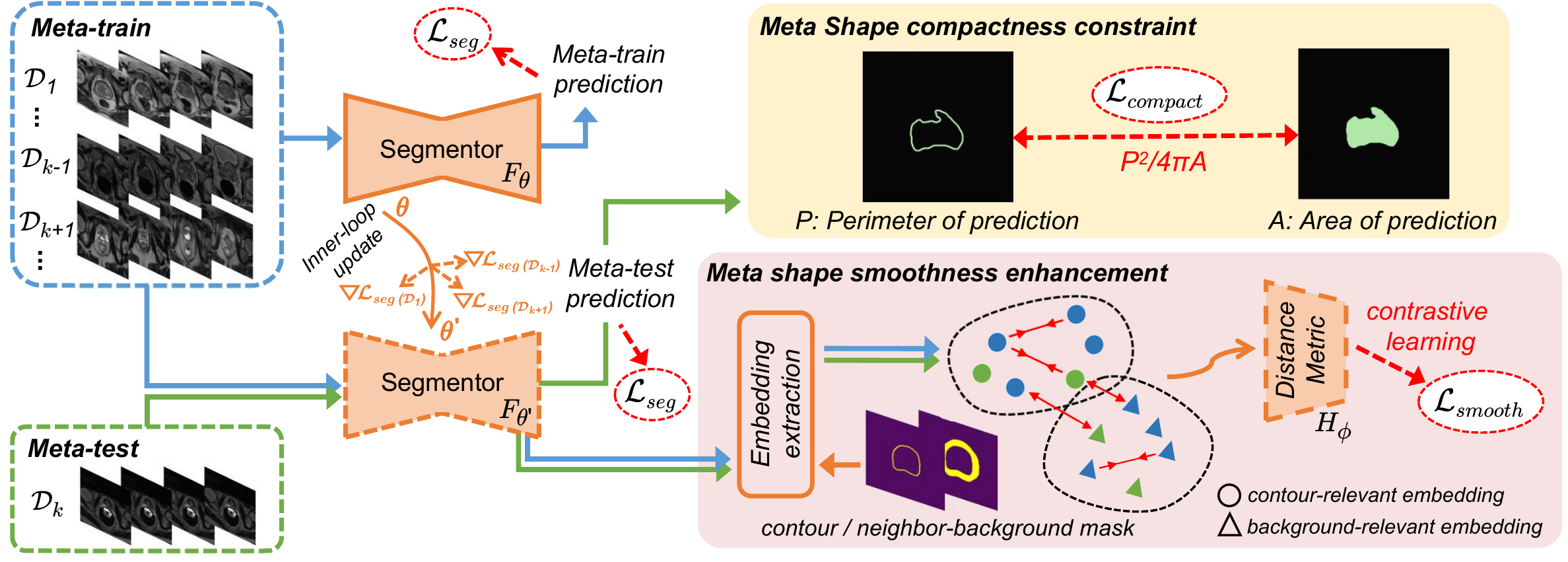}
    	\caption{Overview of our shape-aware meta-learning scheme. The source domains are randomly split into meta-train and meta-test to simulate the domain shift (Sec.~\ref{section:meta}). In meta-optimization: (1) we constrain the shape compactness in meta-test to encourage segmentations with complete shape (Sec.~\ref{section:compact}); (2) we promote the intra-class cohesion and inter-class separation between the contour and background embeddings regardless of domains, to enhance domain-invariance for robust boundary delineation (Sec.~\ref{section:smooth}).}
	\label{fig:overview}
\end{figure*}

Let $(\mathcal{X}, \mathcal{Y})$ denote the joint input and label space in an segmentation task, $\mathcal{D}=\{\mathcal{D}_1,\mathcal{D}_2,...,\mathcal{D}_K\}$ be the set of $K$ source domains. Each domain $\mathcal{D}_k$ contains image-label pairs $\{(x^{(k)}_n,y^{(k)}_n)\}_{n=1}^{N_k}$ sampled from domain distributions $(\mathcal{X}_k, \mathcal{Y})$, where $N_k$ is the number of samples in the $k$-th domain. Our goal is to learn a segmentation model $F_\theta:\mathcal{X} \! \rightarrow \! \mathcal{Y}$ using all source domains $\mathcal{D}$ in a way such that it generalizes well to an unseen target domain $\mathcal{D}_{tg}$. Fig.~\ref{fig:overview} gives an overview of our proposed shape-aware meta-learning scheme, which we will detail in this section.

\subsection{Gradient-based Meta-learning Scheme}
\label{section:meta}
The foundation of our learning scheme is the gradient-based meta-learning algorithm~\cite{li2018learning}, to promote robust optimization by simulating the real-world domain shifts in the training process. 
Specifically, at each iteration, the source domains $\mathcal{D}$ are randomly split into the meta-train $\mathcal{D}_{tr}$ and meta-test $\mathcal{D}_{te}$ sets of domains. The meta-learning can be divided into two steps. 
First, the model parameters $\theta$ are updated on data from meta-train $\mathcal{D}_{tr}$, using Dice segmentation loss $\mathcal{L}_{seg}$:
\begin{equation}
\theta' = \theta - \alpha \nabla_{\theta}\mathcal{L}_{seg}(\mathcal{D}_{tr}; \theta),
\end{equation}
where $\alpha$ is the learning-rate for this inner-loop update. 
Second, we apply a meta-learning step, aiming to enforce the learning on meta-train $\mathcal{D}_{tr}$ to further exhibit certain properties that we desire on unseen meta-test $\mathcal{D}_{te}$. Crucially, the meta-objective $\mathcal{L}_{meta}$ to quantify these properties is computed with the updated parameters $\theta'$, but optimized towards the original parameters $\theta$. 
Intuitively, besides learning the segmentation task on meta-train $\mathcal{D}_{tr}$, such a training scheme further learns how to generalize at the simulated domain shift across meta-train $\mathcal{D}_{tr}$ and meta-test $\mathcal{D}_{te}$. In other words, the model is optimized such that the parameter updates learned on virtual source domains $\mathcal{D}_{tr}$ also improve the performance on the virtual target domains $\mathcal{D}_{te}$, regarding certain aspects in $\mathcal{L}_{meta}$.

In segmentation problems, we expect the model to well preserve the complete shape (compactness) and smooth boundary (smoothness) of the segmentations in unseen target domains. 
To achieve this, apart from the traditional segmentation loss $\mathcal{L}_{seg}$, we further introduce two complementary loss terms into our meta-objective, $\mathcal{L}_{meta} = \mathcal{L}_{seg} + \lambda_1 \mathcal{L}_{compact} + \lambda_2 \mathcal{L}_{smooth}$ ($\lambda_1$ and $\lambda_2$ are the weighting trade-offs), to explicitly impose the shape compactness and shape smoothness of the segmentation maps under domain shift for improving generalization performance.
\subsection{Meta Shape Compactness Constraint}
\label{section:compact}
Traditional segmentation loss functions,~\eg , Dice loss and cross entropy loss, typically evaluate the pixel-wise accuracy, without a global constraint to the segmentation shape. Trained in that way, the model often fails to produce complete segmentations under distribution shift. Previous study have demonstrated that for the compact objects, constraining the shape compactness~\cite{fan2014multiregion} is helpful to promote segmentations for complete shape, as an incomplete segmentation with irregular shape often corresponds to a worse compactness property.

Based on the observation that the prostate region generally presents a compact shape, and such shape prior is independent of observed domains, 
we propose to explicitly incorporate the compact shape constraint in the meta-objective $\mathcal{L}_{meta}$, for encouraging the segmentations to well preserve the shape completeness under domain shift. Specifically, we adopt the well-established Iso-Perimetric Quotient~\cite{li2013efficient} measurement to quantify the shape compactness, whose definition is $C_{IPQ}={4\pi A}/{P^2}$, where $P$ and $A$ are the perimeter and area of the shape, respectively. 
In our case, we define the shape compactness loss as the reciprocal form of this $C_{IPQ}$ metric, and expend it in a pixel-wise manner as follows:
\begin{equation}
\mathcal{L}_{compact}=\frac{P^2}{4\pi A}= \frac{\sum_{i \in \Omega} \sqrt{(\nabla p_{u_{i}}) ^2 + (\nabla p_{v_{i}}) ^2 +\epsilon}}{ 4\pi (\sum_{i \in \Omega} { |p_i| + \epsilon})},
\end{equation}
where $p$ is the prediction probability map, $\Omega$ is the set of all pixels in the map; $\nabla p_{u_{i}}$ and $\nabla p_{v_{i}}$ are the probability gradients for each pixel $i$ in direction of horizontal and vertical; $\epsilon$ ($1e^{-6}$ in our model) is a hyperparameter for computation stability. Overall, the perimeter length $P$ is the sum of gradient magnitude over all pixels $i\in \Omega$; the area $A$ is calculated as the sum of absolute value of map $p$.  

Intuitively, minimizing this objective function encourages segmentation maps with complete shape, because an incomplete segmentation with irregular shape often presents a relatively smaller area $A$ and relatively larger length $P$, leading to a higher loss value of $\mathcal{L}_{compact}$. Also note that we only impose $\mathcal{L}_{compact}$ in meta-test $\mathcal{D}_{te}$, as we expect the model to preserve the complete shape on unseen target images, rather than overfit the source data.

\subsection{Meta Shape Smoothness Enhancement}
\label{section:smooth}
In addition to promoting the complete segmentation shape, we further encourage smooth boundary delineation in unseen domains, by regularizing the model to capture domain-invariant contour-relevant and background-relevant embeddings that cluster regardless of domains. This is crucial, given the observation that performance drop at the cross-domain deployment mainly comes from the ambiguous boundary regions.
In this regard, we propose a novel objective $\mathcal{L}_{smooth}$ to enhance the boundary delineation, by explicitly promoting the intra-class cohesion and inter-class separation between the contour-relevant and background-relevant embeddings drawn from each sample across all domains $\mathcal{D}$.

Specifically, given an image $x_m \in \mathbb{R}^{H\times W\times 3}$ and its one-hot label $y_m$, we denote its activation map from layer $l$ as $M^l_m \in \mathbb{R}^{H_l \times W_l \times C_l}$, and we interpolate $M_m^l$ into $T_m^l \in \mathbb{R}^{H \times W \times C_l}$ using bilinear interpolation to keep consistency with the dimensions of $y_m$. To extract the contour-relevant embeddings $E_m^{con} \in \mathbb{R}^{C_l}$ and background-relevant embeddings $E_m^{bg} \in \mathbb{R}^{C_l}$, we first obtain the binary contour mask $c_m \in \mathbb{R}^{H \times W \times 1}$ and binary background mask $b_m \in \mathbb{R}^{H \times W \times 1
}$ from $y_m$ using morphological operation. Note that the mask $b_m$ only samples background pixels around the boundary, since we expect to enhance the discriminativeness for pixels around boundary region. Then, the embeddings $E_m^{con}$ and $E_m^{bg}$ can be extracted from $T_m^l$ by conducting 
weighted average operation over $c_m$ and $b_m$:

\begin{equation}
E_m^{con} = \frac{\sum_{i \in \Omega} (T_m^l)_i \cdot (c_m)_i}{\sum_{i \in \Omega} (c_m)_i}, \quad
E_m^{bg} = \frac{\sum_{i \in \Omega} (T_m^l)_i \cdot (b_m)_i}{\sum_{i \in \Omega} (b_m)_i},
\end{equation}
where $\Omega$ denotes the set of all pixels in $T_m^l$, the $E_m^{con}$ and $E_m^{bg}$ are single vectors, representing the contour and backgound-relevant representations extracted from the whole image $x_m$.
In our implementation, activations from the last two deconvolutional layers are interpolated and concatenated to obtain the embeddings. 


Next, we enhance the domain-invariance of 
$E^{con}$ and $E^{bg}$ in latent space, by encouraging embeddings' intra-class cohesion and inter-class separation among samples from all source domains $\mathcal{D}$. 
Considering that imposing such regularization directly onto the network embeddings might be too strict to impede the convergence of $\mathcal{L}_{seg}$ and $\mathcal{L}_{compact}$, 
we adopt the
contrastive learning~\cite{chen2020simple} to achieve this constraint. Specifically, an embedding network $H_\phi$ is introduced to project the features $E \in [E^{con}, E^{bg} ]$ to a lower-dimensional space, then the distance is computed on the obtained feature vectors from network $H_\phi$ as
$
d_{\phi}(E_m, E_n) = \|H_\phi(E_m) - H_\phi(E_n)\|_2
\label{equ:distance}
$, where the sample pair $(m, n)$ are randomly drawn from all domains $\mathcal{D}$, as we expect to harmonize the embeddings space of $\mathcal{D}_{te}$ and $\mathcal{D}_{tr}$ to capture domain-invariant representations around the boundary region.
Therefore in our model, the contrastive loss is defined as follows:  
\begin{equation}
\ell_{contrastive}(m, n) = 
\left\{
\begin{array}{lr}
d_{\phi}(E_{m}, E_{n}), & ~\text{if} ~ \tau(E_m) = \tau(E_n)\\
(max\{0, \zeta-d_{\phi}(E_{m}, E_{n}\})^2,&  ~\text{if} ~ \tau(E_m) \neq \tau(E_n)\\
\end{array}
\right..
\end{equation}
where the function $\tau(E)$ indicates the class (1 for $E$ being $E^{con}$, and 0 for $E^{bg}$)
$\zeta$ is a pre-defined distance margin following the practice of metric learning (set as 10 in our model). 
The final objective $\mathcal{L}_{smooth}$ is computed within mini-batch of $q$ samples. We randomly employ either $E^{con}$ or $E^{bg}$ for each sample, and the $\mathcal{L}_{smooth}$ is the average of $\ell_{contrastive}$ over all pairs of $(m,n)$ embeddings:
\begin{equation}
\mathcal{L}_{smooth} = \sum\nolimits_{m = 1}^q \sum\nolimits_{n = m +1}^q \ell_{contrastive}(m,n)/ C(q,2).
\end{equation}
where $C(q,2)$ is the number of combinations. Overall, all training objectives 
including $\mathcal{L}_{seg}(\mathcal{D}_{tr}; \theta)$ and $\mathcal{L}_{meta}(\mathcal{D}_{tr},D_{te}; \theta')$, are optimized together with respect to the original parameters $\theta$. The $\mathcal{L}_{smooth}$ is also optimized with respect to $H_\phi$.

\section{Experiments}
\begin{table}[b]
    \renewcommand\arraystretch{1.1}
    \centering
    \caption{Details of our employed six different sites obtained from public datasets.}
    \label{table:dataset}
    \scalebox{0.73}{
    \begin{tabular}{p{1.5cm}|p{1.5cm}|p{1.5cm}|p{1.8cm}|p{3.0cm}|p{1.8cm}|p{1.5cm}}
    \hline
    Dataset &Institution  & Case num     & Field strength(T) & Resolution(in/ through plane)(mm) & Endorectal Coil & Manufactor \\
    \hline
    Site A &RUNMC & 30  & 3 & 0.6-0.625/3.6-4 & Surface & Siemens  \\
    Site B &BMC & 30  & 1.5& 0.4/3 & Endorectal& Philips   \\
    Site C &HCRUDB & 19  & 3 & 0.67-0.79/1.25   & No & Siemens \\
    Site D &UCL & 13  & 1.5 and 3 & 0.325-0.625/3-3.6   & No & Siemens \\
    Site E &BIDMC & 12  & 3 & 0.25/2.2-3   & Endorectal & GE \\
    Site F &HK  & 12  & 1.5 & 0.625/3.6   & Endorectal & Siemens \\
    \hline
    \end{tabular}
    }
\end{table}
\subsubsection{Dataset and Evaluation Metric.}
We employ prostate T2-weighted MRI from 6 different data sources with distribution shift (cf. Table~\ref{table:dataset} for summary of their sample numbers and scanning protocols). 
Among these data, samples of Site A,B are from NCI-ISBI13 dataset~\cite{bloch2015nci}; samples of Site C are from I2CVB dataset~\cite{lemaitre2015computer}; samples of Site D,E,F are from PROMISE12 dataset~\cite{litjens2014evaluation}.
Note that the NCI-ISBI13 and PROMISE12 actually include multiple data sources, hence we decompose them in our work. 
For pre-processing, we resized each sample to $384 \! \times \! 384$ in axial plane, and normalized it to zero mean and unit variance. We then clip each sample to only preserve slices of prostate region for consistent objective segmentation regions across sites. We adopt Dice score (Dice) and Average Surface Distance (ASD) as the evaluation metric.
\subsubsection{Implementation Details. }
We implement an adapted Mix-residual-UNet~\cite{yu2017volumetric} as segmentation backbone. Due to the large variance on 
slice thickness among different sites, we employ the 2D architecture. 
The domains number of meta-train and meta-test were set as 2 and 1.
The weights $\lambda_1$ and $\lambda_2$ were set as 1.0 and $5e^{-3}$. The embedding network $H_\phi$ composes of two fully connected layers with output sizes of 48 and 32. 
The segmentation network $F_\theta$ was trained using Adam optimizer and the learning rates for inner-loop update and meta optimization were both set as $1e^{-4}$. The network $H_\phi$ was also trained using Adam optimizer with learning rate of $1e^{-4}$.  We trained 20K iterations  with batch size of 5 for each source domain. For batch normalization layer, we use the statistics of testing data for feature normalization during inference for better generalization performance.
\subsubsection{Comparison with State-of-the-art Generalization Methods. }
We implemented several state-of-the-art generalization methods for comparison, including a data-augmentation based method (BigAug)~\cite{zhang2020generalizing}, a classifier regularization based method (Epi-FCR)~\cite{li2019episodic}, a latent space regularization method (LatReg)~\cite{aslani2020scanner} and a meta-learning based method (MASF)~\cite{dou2019domain}. 
In addition, we conducted experiments with `DeepAll' baseline (i.e., aggregating data from all source domains for training a deep model) and `Intra-site' setting (i.e., training and testing on the same domain, with some outlier cases excluded to provide general internal performance on each site). 
Following previous practice~\cite{dou2019domain} for domain generalization,
we adopt the leave-one-domain-out strategy,~\ie, training on $K$-1 domains and testing on the one left-out unseen target domain.
    
\begin{figure}[t]
	\centering
	\includegraphics[width=0.9\textwidth]{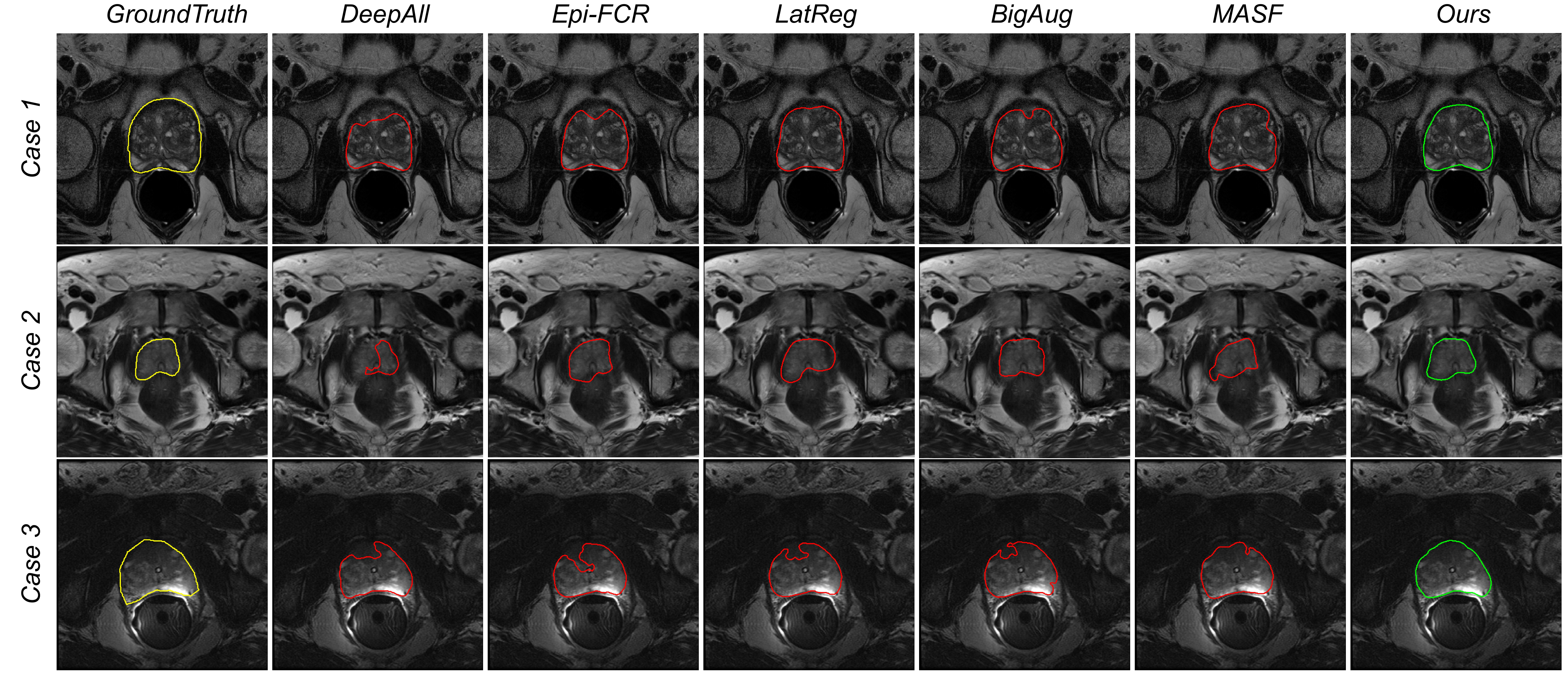}
	\caption{Qualitative comparison on the generalization results of different methods, with three cases respectively drawn from different unseen domains.} 
	\label{fig:qualitive_result}
\end{figure}

As listed in Table~\ref{table:results}, 
DeepAll presents a strong performance, while the Epi-FCR with classifier regularization shows limited advantage over this baseline. 
The other approaches of LatReg, BigAug and MASF are more significantly better than DeepAll, with the meta-learning based method yielding the best results among them in our experiments.
Notably, our approach (cf. the last row) achieves higher performance over all these state-of-the-art methods across all the six sites,
and outperforms the DeepAll model by 2.15\% on Dice and 0.60$mm$ on ASD, demonstrating the capability of our shape-aware meta-learning scheme to deal with domain generalization problem.
Moreover, Fig.~\ref{fig:qualitive_result} shows the generalization segmentation results of different methods on three typical cases from different unseen sites. We observe that 
our model with shape-relevant meta regularizers can well preserve the complete shape and smooth boundary for the segmentation in unseen domains, whereas other methods sometimes failed to do so.
\begin{table}[t]
    \renewcommand\arraystretch{1.1}
    \centering
    \caption{Generalization performance of various methods on Dice (\%) and ASD ($mm$).}
    \label{table:results}
    \scalebox{0.73}{
    \begin{tabular}{p{2.8cm}|cc|cc|cc|cc|cc|cc|cc}
    \hline
    Method &  \multicolumn{2}{c|}{Site A} & \multicolumn{2}{c|}{Site B} & \multicolumn{2}{c|}{Site C} &\multicolumn{2}{c|}{Site D} &\multicolumn{2}{c|}{Site E} &\multicolumn{2}{c|}{Site F} &\multicolumn{2}{c}{Average}\\
    \hline

Intra-site &89.27 &1.41 &\underline{88.17} &1.35 &\underline{88.29} &1.56 &83.23 &3.21 &83.67 &2.93 &85.43 &1.91 &86.34 &2.06 \\
    \hline
    DeepAll (baseline) &87.87  &2.05 &85.37 &1.82 &82.94 &2.97 &86.87 &2.25 &84.48 &2.18 &85.58 &1.82 &85.52 &2.18\\
    \hline
    Epi-FCR~\cite{li2019episodic} &88.35 &1.97 &85.83 &1.73 &82.56 &2.99 &86.97 &2.05 &85.03 &1.89 &85.66 &1.76 &85.74  &2.07\\
    LatReg~\cite{aslani2020scanner} &88.17 &1.95 &86.65 &1.53 &83.37 &2.91 &87.27 &2.12 &84.68 &1.93 &86.28 &1.65 &86.07 &2.01\\
    BigAug~\cite{zhang2020generalizing}&88.62 &1.70 &86.22 &1.56 &83.76 &2.72 &87.35 &1.98 &85.53 &1.90 &85.83 &1.75 &86.21 &1.93\\
    
    MASF~\cite{dou2019domain} &88.70 &1.69 &86.20 &1.54 &84.16 &2.39 &87.43 &1.91 & 86.18 &1.85 &86.57 &1.47 &86.55 &1.81\\
    \hline
    Plain meta-learning &88.55 &1.87 &85.92 &1.61 &83.60 &2.52 &87.52 &1.86 &85.39 &1.89 &86.49 &1.63 &86.24 &1.90\\
    + $\mathcal{L}_{compact}$ &89.08 &1.61 &87.11 &1.49 &84.02 &2.47 &87.96 &1.64 &86.23 &1.80 &87.19 &1.32 &86.93 &1.72\\
    + $\mathcal{L}_{smooth}$ &89.25 &1.64&87.14 &1.53 &~\textbf{84.69}&2.17 &87.79&1.88&86.00 &1.82&87.74 &1.24 &87.10 &1.71\\
    SAML (\textbf{Ours}) &~\textbf{89.66} &~\textbf{1.38} &~\textbf{87.53} &~\textbf{1.46} &84.43 &~\textbf{2.07} &~\textbf{88.67}  &~\textbf{1.56} &~\textbf{87.37} &~\textbf{1.77} &~\textbf{88.34} &~\textbf{1.22} &~\textbf{87.67} &~\textbf{1.58}\\
    \hline
    \end{tabular}
    }
\end{table}
We also report in Table~\ref{table:results} the cross-validation results conducted within each site, i.e., Intra-site.
Interestingly, we find that this result for site D/E/F is relatively lower than the other sites, and even worse than the baseline model. The reason would be that the sample numbers of these three sites are fewer than the others, consequently intra-site training is ineffective with limited generalization capability.
This observation reveals the important fact that, when a certain site suffers from severe data scarcity for model training, aggregating data from other sites (even with distribution shift) can be very helpful to obtain a qualified model.
In addition, we also find that our method outperforms the Intra-site model in 4 out of 6 data sites, with superior overall performances on both Dice and ASD, which endorses the potential value of our approach in clinical practice.
\begin{figure*}[t]
	\centering
	\includegraphics[width=0.93\textwidth]{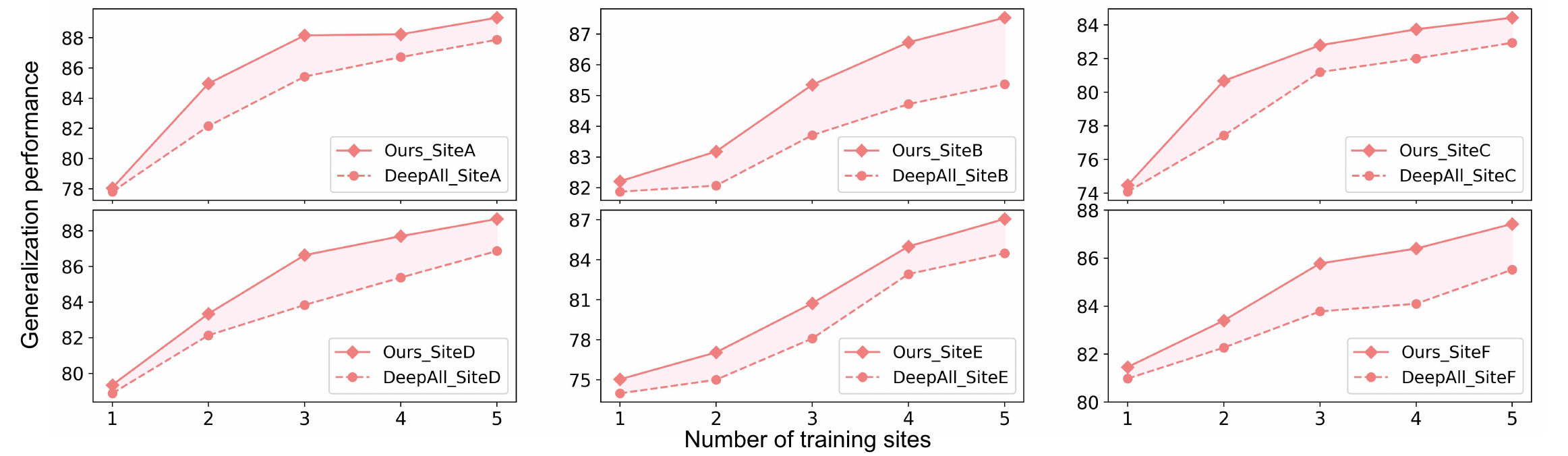}
	\caption{Curves of generalization performance on unseen domain as the number of training source domain increases, using DeepAll method and our proposed approach.} 
	\label{fig:domain_number}
\end{figure*}
\subsubsection{Ablation Analysis. }
We first study the contribution of each key component in our model. 
As shown in Table~\ref{table:results},
the plain meta-learning method only with $\mathcal{L}_{seg}$ can already outperform the DeepAll baseline, leveraging the explicit simulation of domain shift for training.
Adding shape compactness constraint into $\mathcal{L}_{meta}$ yields improved Dice and ASD which are higher than MASF. 
Further incorporating $L_{smooth}$ (SAML) to encourage domain-invariant embeddings for pixels around the boundary, 
consistent performance improvements on all six sites are attained. Besides, simply constraining $L_{smooth}$ on pure meta-learning method (+ $L_{smooth}$) also leads to improvements across sites.

We further investigate the influence of training domain numbers on the generalization performance of our approach and the DeepAll model. Fig.~\ref{fig:domain_number} illustrates how the segmentation performance on each unseen domain would change, as we gradually increase the number of source domains in range $[1, K\!-\!1]$.
Obviously, when a model is trained just with a single source domain, directly applying it to target domain receives unsatisfactory results.
The generalization performance progresses as the training site number increases, indicating that aggregating wider data sources helps to cover a more comprehensive distribution. 
Notably, our approach consistently outperforms DeepAll across all numbers of training sites, confirming the stable efficacy of our proposed learning scheme. 

\section{Conclusion}
We present a novel shape-aware meta-learning scheme to improve the model generalization in prostate MRI segmentation. On top of the meta-learning strategy, we introduce two complementary objectives to enhance the segmentation outputs on unseen domain by imposing the shape compactness and smoothness in meta-optimization. Extensive experiments demonstrate the effectiveness. To our best knowledge, this is the first work incorporating shape constraints with meta-learning for domain generalization in medical image segmentation. Our method can be  extended to various segmentation scenarios that suffer from domain shift.
\subsubsection{Acknowledgement.} This work was supported in parts by the following grants:
Key-Area Research and Development Program of Guangdong Province, China (2020B010165004),
Hong Kong Innovation and Technology Fund (Project No. ITS/426/17FP),
Hong Kong RGC TRS Project T42-409/18-R,  and National Natural Science 
Foundation of China with Project No. U1813204.
\bibliographystyle{splncs04}
\bibliography{refs}

\end{document}